\documentclass[runningheads]{llncs}
\usepackage{eccv}
\usepackage{eccvabbrv}
\usepackage{graphicx}
\usepackage{booktabs}
\usepackage[accsupp]{axessibility}
\usepackage[pagebackref,breaklinks,colorlinks]{hyperref}
\usepackage{orcidlink}
\usepackage{amsmath, amssymb, bbm} 
\usepackage{algorithm}
\usepackage{algorithmic}
\usepackage{graphicx}
\usepackage{caption}
\usepackage{subcaption}
\usepackage{colortbl}
\usepackage{listings}
\usepackage{bbding}
\usepackage[misc]{ifsym}
\usepackage{tikz}

\usetikzlibrary{fit,calc}

\colorlet{algogreen}{red!40}
\colorlet{algoblue}{cyan!60}
\colorlet{algogreen}{green!60}
\definecolor{dkgreen}{rgb}{0,0.6,0}

\begin{document}

\title{Diffusion-based Aesthetic QR Code Generation via Scanning-Robust Perceptual Guidance}

\titlerunning{Diffusion-based Aesthetic QR Code Generation}

\author{
Jia-Wei Liao \inst{1, 2} \orcidlink{0000-0001-7328-0961}
\and
Winston Wang \inst{1}$^\star$
\and
Tzu-Sian Wang \inst{1}$^\star$
\and
Li-Xuan Peng \inst{1}\thanks{Equal contribution}
\and \\
Cheng-Fu Chou \inst{2} \orcidlink{0000-0003-2684-5039} \and 
Jun-Cheng Chen \inst{1}
\textsuperscript{\Letter}
\orcidlink{0000-0002-0209-8932}
}

\authorrunning{J.-W. Liao et al.}
\institute{
Research Center for Information Technology Innovation, Academia Sinica \and
National Taiwan University \\
\email{
\{d11922016, ccf\}@csie.ntu.edu.tw, \\
\{winston0724, as6325400, alexpeng517\}@gmail.com, \\
pullpull@citi.sinica.edu.tw
}
}

\maketitle

\begin{figure}[h]
    \centering
    \includegraphics[height=3.5cm]{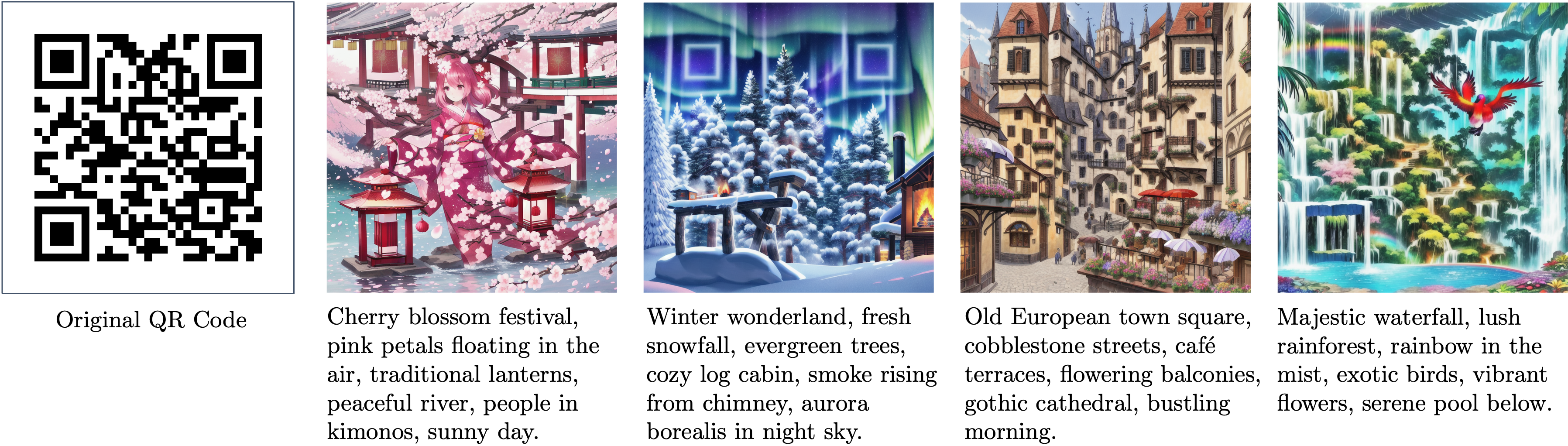}
    \caption{Leveraging the preeminent capability of Latent Diffusion Model (LDM) and ControlNet as a prior knowledge of aesthetic QR code images, coupled with our proposed Scanning-Robust (Perceptual) Guidance, we can generate custom-styled QR codes conform to user prompts while assuring both scannability and aesthetics.}
    \label{fig:teaser}
    \vspace{-30pt}
\end{figure}

\begin{abstract}
QR codes, prevalent in daily applications, lack visual appeal due to their conventional black-and-white design. Integrating aesthetics while maintaining scannability poses a challenge. In this paper, we introduce a novel diffusion-model-based aesthetic QR code generation pipeline, utilizing pre-trained ControlNet and guided iterative refinement via a novel classifier guidance (SRG) based on the proposed Scanning-Robust Loss (SRL) tailored with QR code mechanisms, which ensures both aesthetics and scannability. To further improve the scannability while preserving aesthetics, we propose a two-stage pipeline with Scanning-Robust Perceptual Guidance (SRPG). Moreover, we can further enhance the scannability of the generated QR code by postprocessing it through the proposed Scanning-Robust Projected Gradient Descent (SRPGD) post-processing technique based on SRL with proven convergence. With extensive quantitative, qualitative, and subjective experiments, the results demonstrate that the proposed approach can generate diverse aesthetic QR codes with flexibility in detail. In addition, our pipelines outperforming existing models in terms of Scanning Success Rate (SSR) 86.67\% (+40\%) with comparable aesthetic scores. The pipeline combined with SRPGD further achieves 96.67\% (+50\%). Our code will be available \url{https://github.com/jwliao1209/DiffQRCode}.

\end{abstract}

\section{Introduction}

\begin{figure}[t]
    \centering
    \includegraphics[height=2.4cm]{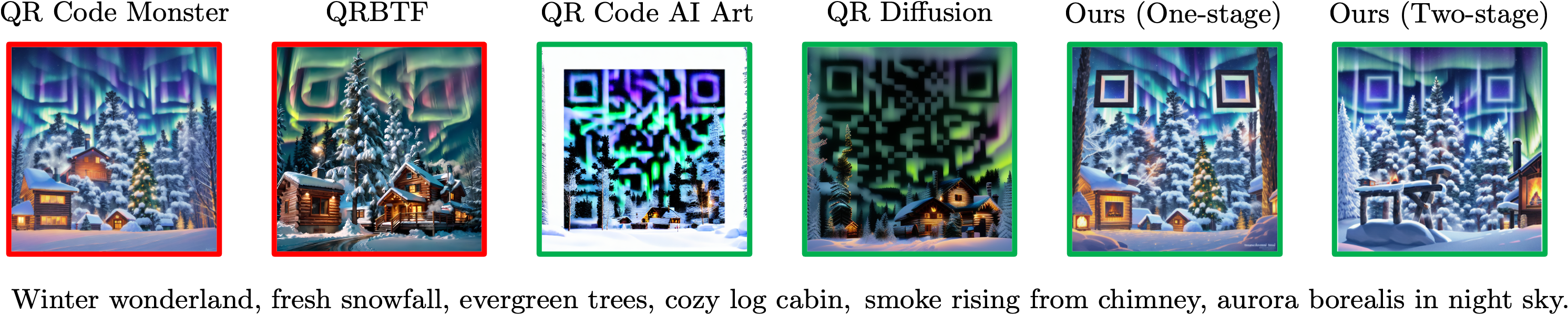} 
    \caption{The scannability and aesthetics dilemma of prevailing methods. QR Code Monster\cite{qrcodemonster2023} and QRBTF \cite{qrbtf2023} are capable of generating visual-appealing QR codes but the scannability is uncertain; QR Code AI Art \cite{qrcodeai} and QR Diffusion \cite{qrdiffusion} are capable of generating scannable QR codes but with limited aesthetics; Our proposed one-stage generation pipeline could generate both aesthetic and scanning-robust QR codes, and our proposed two-stage generation pipeline further improved the visual quality by harmonically merging the QR code alignment pattern into prompt-specific semantics. Red frames indicated unscannable, while green frames indicated scannable, zoom in for better details.}
    \label{fig:dilemma}
    \vspace{-10pt}
\end{figure}

Quick Response (QR) codes have become ubiquitous in daily applications in recent years due to their rapid, efficient readability and widespread availability of smartphone QR code scanners. These codes are widely adopted in various scenarios, such as daily transactions, spreading information, interactive marketing, and more. However, conventional QR codes usually comprise semantically inconspicuous black-and-white pixels, lacking visual appeal. Integrating designs or textures into QR codes can enhance their attractiveness, capturing the users' attention. Recognizing the commercial value of aesthetically pleasing QR codes, researchers have developed numerous integration techniques, seamlessly incorporating QR codes into branding and marketing materials.

One major challenge is to fuse natural, semantically meaningful reference images and QR codes, endowing the aesthetics towards conventional QR codes. Previous research focused on techniques based on combining QR code modules and style transfer. These methods allow embedding QR code patterns into style textures, producing visually appealing QR codes, but still possessing limited flexibility and fine-grained editability.

Integrating QR code patterns with semantically meaningful reference images without relying on style-transfer-based methods poses a challenge, and sub-optimal integration could compromise both scannability and aesthetics. Consequently, prevailing commercial products adopted generative models to create aesthetic QR codes, primarily employing diffusion models and ControlNet \cite{qrcodemonster2023}. The mainstream methodology is adjusting the ControlNet classifier-free guidance (CFG) weight to generate aesthetic QR codes. However, balancing the CFG weight forms a dilemma (Fig.~\ref{fig:dilemma}): lower weight on guidance inhibits QR code scannability while higher guidance compromises aesthetics. 
In practical schemes, it's common to leverage manual post-processing, but it is time-consuming and labor-intensive. Our work seeks to address this issue, i.e., effectively balance the scannability and aesthetics.

In this paper, we introduce diffusion-model-based aesthetic QR code generation pipelines with iterative refinement via our proposed Scanning-Robust Guidance (SRG). More specifically, our proposed one-stage pipeline utilizes pre-trained ControlNet conditioned on conventional QR codes and user prompts, followed by our SRG to iteratively refine the fine-grained semantics of intermediate images to assure scannability. The key insight of our SRG design is to leverage the conventional QR code pixel redundancy and error tolerance property to devise a novel soft-decision Scanning-Robust Loss (SRL) as the generalized conditional probability guidance. Moreover, our SRL has an early-stopping mechanism to avoid over-optimization-induced artifacts centered on modules. To further improve the performance, we respectively propose the two-stage generation pipeline with additional perceptual constraints on top of SRG and the post-processing technique via Scanning-Robust Projected Gradient Descent (SRPGD). The former can enhance aesthetics while preserving scannability, and the latter ensures near-perfect scannability with little visual quality degeneration via convex optimization. 

Compared to pre-existing style transfer methods, our approach can generate more diverse aesthetic QR codes with larger flexibility in details. Furthermore, extensive experimental results demonstrate the effectiveness of our approaches while outperforming existing open-source and proprietary generative models in terms of Scanning Success Rate (SSR) and LAION Aesthetics Score (LAS) $^{\color{red}3}$, both quantitatively and qualitatively.
Specifically, our one-stage pipeline achieves 83.33\% SSR (+36.66\%) with a minor LAS decrease; our two-stage pipeline achieves 86.67\% SSR (+40\%) while better preserving the LAS score. Combined with SRPGD, our one-stage pipeline achieves 100\% SSR (+53.33\%); our two-stage pipeline achieves 96.67\% SSR (+50\%), both with little LAS score decreases. Some of our generation results are also illustrated in Fig.~\ref{fig:teaser}.

Our main contributions are summarized as follows:
\begin{enumerate}
    \item We propose a novel diffusion-based iterative refinement with Scanning-Robust Guidance (SRG) tailored with QR code mechanisms.
    \item We propose a two-stage pipeline with Scanning-Robust Perceptual Guidance (SRPG) to improve aesthetics further while preserving scannability.
    \item We propose a post-processing technique that ensures scannability convergence via the convex property of our proposed Scanning-Robust Loss (SRL).
\end{enumerate}

\section{Related Work}

\subsection{Image Diffusion Models}
Recently, Diffusion Models \cite{sohl2015deep, ho2020denoising} emerged as powerful generative models, showcasing remarkable unconditional image generation capabilities compared to pre-dominant GAN-based models \cite{goodfellow2020generative, dhariwal2021diffusion}. 
To impart semantic controllability to diffusion models, Dhariwal et al. \cite{dhariwal2021diffusion} introduced the classifier guidance, \cite{liu2023more, kim2022diffusionclip, zhao2022egsde, avrahami2022blended, yu2023freedom} further expanded this concept and apply to broader conditional image generation scenarios.

However, diffusion models required more sampling steps, thus eliciting higher computational burdens, especially when dealing with high-resolution images. To address this problem, Rombach et al. \cite{rombach2022high} proposed the Latent Diffusion Model (LDM), transferring the diffusion process from pixel space to lower dimensional latent space via the pre-trained image reconstruction VAEs. For more fine-grained manipulations and adaptations for downstream tasks based on large pre-trained LDM, Zhang et al., Qin et al., and Zavadski et al. \cite{zhang2023adding, qin2023unicontrol, zavadski2023controlnet} proposed fine-tuning only on additional layers instead of fine-tuning as a whole, achieving exceptional task-specific performance with little cost. Numerous studies also focus on reducing the steps needed while maintaining overall sampling quality to accelerate the sampling speed further. 
These advancements have propelled the flourishing in the fields of image editing\cite{meng2021sdedit, dhariwal2021diffusion, nichol2022glide, couairon2022diffedit, hertz2022prompt, yang2024dynamic, mokady2023null}, text-to-image synthesis \cite{rombach2022high, ramesh2022hierarchical, ruiz2023dreambooth}, and commercial products including DALL-E2 \cite{openai2023dalle2} and Midjourney \cite{midjourney2023}
Our work employs this powerful prior knowledge of diffusion models as our aesthetic QR code generation backbone.

\subsection{Aesthetic QR Codes}
Previous aesthetic QR code research has mainly focused on QR code module-based and style transfer techniques.
Module-based techniques are mainly referred to as module-deformation and module-reshuffle. Module-deformation combines reference image and QR code via deforming and scaling operations on the QR code module. Visualead \cite{visualead}, LogoQ \cite{logoq}, and Halftone QR code. Chu et al. \cite{chu2013halftone} proposed blending the reference image with the area of each module other than the center pixels to ensure scannability and aesthetics. Module-reshuffle, first proposed by Qart \cite{qartcodes}, leverages the Gaussian-Jordan elimination procedure to reshuffle the QR code modules, aiming to allow the distribution of black-and-white pixels to match the pattern of reference images while ensuring the decoding correctness. Subsequent research based on image processing techniques, e.g., the region of interest \cite{xu2021art}, central saliency \cite{lin2015efficient}, and global gray values \cite{xu2019stylized}, further improved the visual quality of aesthetic QR code.

Xu et al. \cite{xu2019stylized} proposed Stylized aEsthEtic (SEE) QR code, the first work to utilize the style-transfer techniques for aesthetic QR code generation. SEE resolved the scannability degeneration induced by style transfer via the post-processing algorithm. However, it might lead to concentrations of black-and-white pixels centered on QR code modules, engendering unnatural artifacts. 

Su et al.\cite{su2021artcoder} introduced an end-to-end architecture, alleviating SEE artifacts by balancing style characteristics, semantics, and readability. More specifically, they jointly optimize style, content, and code losses. Still, it has room for improvement, as we can find minor artifacts when zooming in. Q-Art Code \cite{su2021q} introduced the Module-based Deformable Convolutional Mechanism (MDCM), letting artifacts of sub-optimal QR code harmonically merge in styled QR code. 

With the growth of diffusion-model-based image manipulation and conditional controlling, various commercial products \cite{qrdiffusion, qrcodeai, openart, huggingface} leverage these generative models to create aesthetic QR code. Nevertheless, current methods are not yet being systematically designed against the inherent mechanisms of QR codes. Consequently, it's difficult to strike a balance between scannability and aesthetics.

\section{Method}
\begin{figure}[t]
    \centering
    \includegraphics[height=6cm]{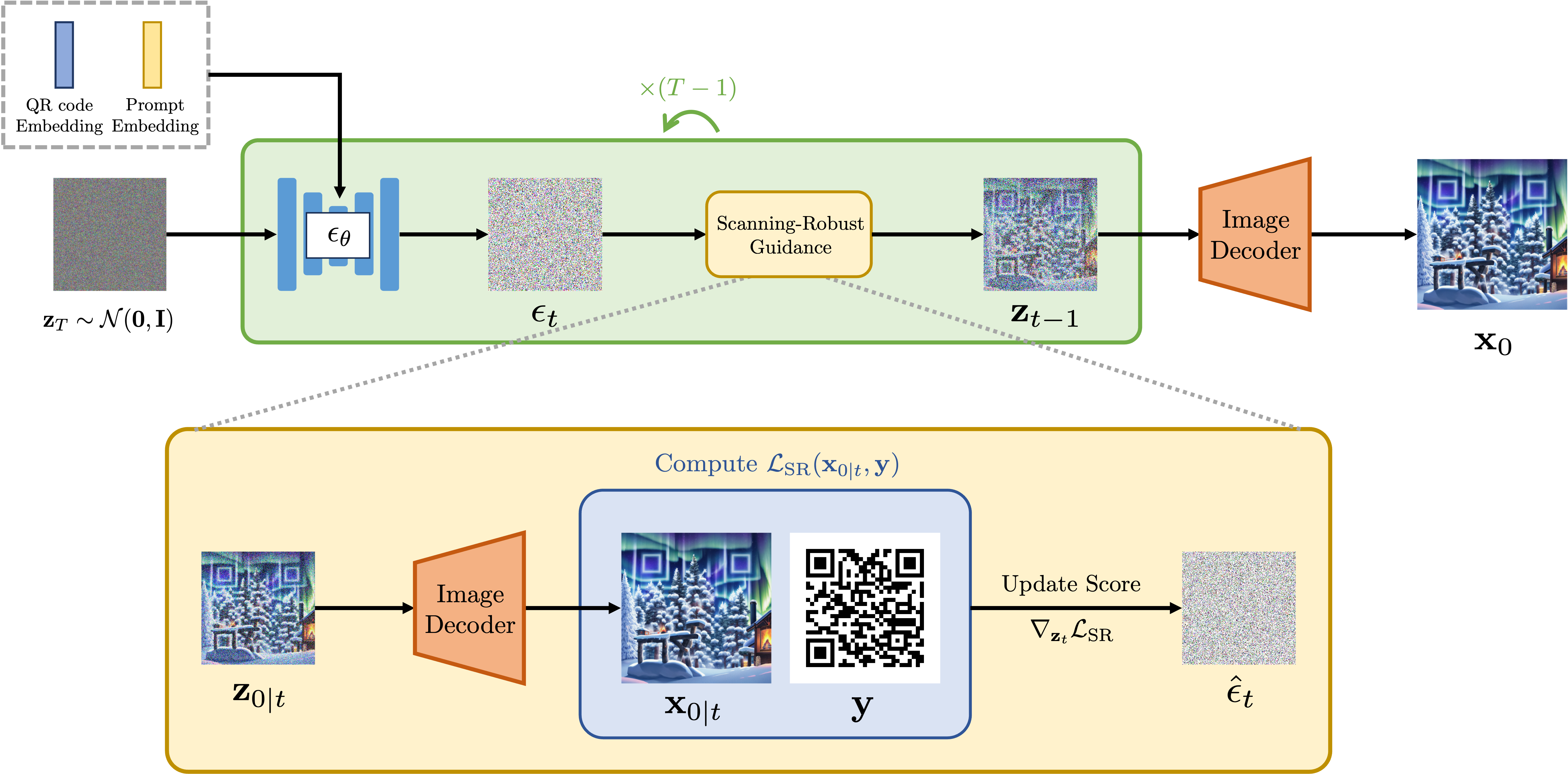}
    \caption{\textbf{An overview of our proposed iterative refinement with Scanning Robustness Guidance (SRG).} First, we leverage pre-trained ControlNet to obtain the initial score prediction conditioned on the target QR code and user-specified prompt. During each denoising step, we approximate $\mathbf{z}_{0|t}$ followed by DDIM formulation, then apply the VAE decoder to get $\mathbf{x}_{0|t}$ for $\mathcal{L}_\text{SR}$ calculation. We utilize the gradient of $\mathcal{L}_\text{SR}$ as a guidance term to update the predicted score. Repeat the above iterative refinement process until convergence.}
    \label{fig:pipeline}
    \vspace{-10pt}
\end{figure}

Our scanning-robust aesthetic QR code generation pipeline comprises two components, pre-trained ControlNet and our proposed guided iterative refinement. We leverage pre-trained ControlNet checkpoints conditioned on the target input QR codes as the backbone for our guided iterative refinement as shown in Fig.~\ref{fig:pipeline}.

Specifically, in Sec. \ref{sec:SRL}, we propose Scanning-Robust Loss (SRL) to evaluate the scannability of the generated aesthetic QR code, effectively via leveraging the inherent redundancy and error correction mechanisms of conventional QR codes to trading-off QR code scannability and aesthetics. Sec. \ref{sec:OneStage} extends the conditional probability concept to propose the Scanning-Robust Guidance (SRG) integrating with Scanning-Robust Loss (SRL) for our one-stage generation pipeline with iterative refinement. To further improve the aesthetics and scannability. Sec. \ref{sec:TwoStage} proposes the two-stage generation pipeline with additional perceptual constraints. Finally, if the scannability is prioritized, Sec. \ref{sec:SRPGD} introduces the post-processing technique via Scanning-Robust Projected Gradient Descent (SRPGD). By the convex property of our SRL, we can ensure the scannability convergence.

\subsection{Scanning-Robust Loss}
\label{sec:SRL}
\begin{figure}[t]
    \centering
    \includegraphics[height=4cm]{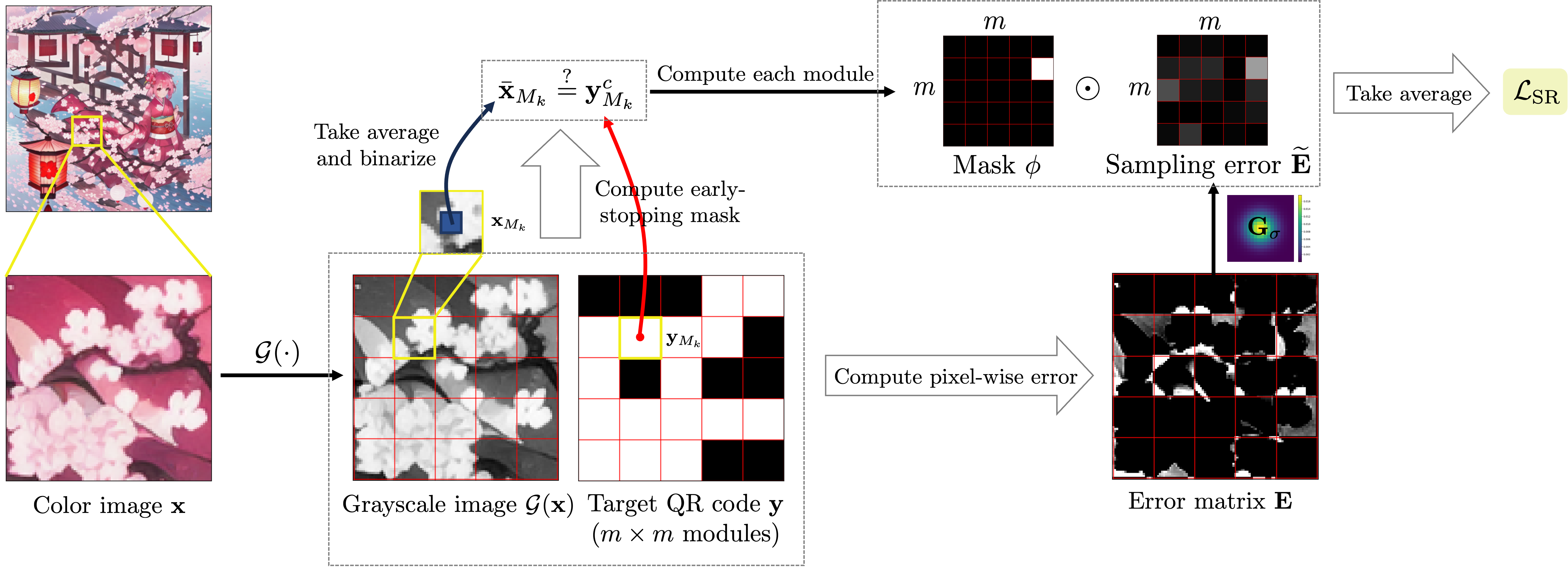}
    \caption{\textbf{An illustration of our proposed Scanning-Robust Loss (SRL).} Without losing the generality, here we demonstrate a small region. We emulate the scanning process using module pixel extraction and binarization to calculate the pixel-wise error matrix and module-wise optimization decision mask. Then we apply a Gaussian kernel to re-weight the error matrix. Finally, we mask the error matrix with the decision mask via Hadamard product, then take the average to form our SRL.}
    \label{fig:srl}
    \vspace{-10pt}
\end{figure}

The proposed Scanning-Robust Loss (SRL) (Fig.~\ref{fig:srl}) is differentiable and capable of measuring the scannability in a soft-decision way. Since the SRL is defined in pixel space, we decode the latent codes back to pixel space before computing it.

\vspace{-10pt}

\subsubsection{Pixel-wise Error.}
Given a normalized image $\mathbf{x}$ \footnote{We normalized all the pixel of images to $[0, 1]$.}, a QR code image $\mathbf{y}$ and gray-scale conversion operator $\mathcal{G}(\cdot)$. We devise the loss to be proportional to the deviation corresponding to the target QR code pixel from the standard binarized threshold $\frac{1}{2}$. We formulate the distance between the aesthetic QR code and the target QR code as linear functions. Combining black and white pixel cases, we can obtain error matrix $\mathbf{E}$ as follows:
\begin{align}
    \label{eq:sr_error}
    \mathbf{E}=
    \max(1-2\mathcal{G}(\mathbf{x}), 0)
    \odot
    \mathbf{y} +
    \max(2\mathcal{G}(\mathbf{x})-1, 0)
    \odot
    (1-\mathbf{y}),
\end{align}
\noindent where $\max(\cdot, \cdot)$ is applied component-wisely, and $\odot$ is the Hadamard product.

\vspace{-10pt}

\subsubsection{Error Re-weighting by Gaussian Kernel.}
Ideally, when $\mathbf{E}$ is optimized to zero matrix $\mathbf{O}$, binarized $\mathbf{x}$ and $\mathbf{y}$
will be identical, indicating the absolute scannable QR code generation.
However, a caveat is that our aesthetic QR code will share excessive structural detail with the target QR code, resulting in inadequate capability for our aesthetic purposes.
To address these problems, we introduce the module concept from QR codes, allowing more extent of aesthetic modifications.
ZXing \cite{ZXing} is an often-used QR code open-sourced project. It points out that only the pixel centered in the module matters, whereas others are negligible for overall scanning correctness.
According to \cite{xu2021art,su2021artcoder}, QR code is sensitive regarding scanning devices and environmental factors, e.g., distortion induced by capture angle and lighting conditions, thus they assume the probability of pixels recognized by devices as representative of that module follows a two-dimensional Gaussian distribution centered on the module. Therefore, we apply a Gaussian kernel function $\boldsymbol{G}_\sigma$ with standard deviation $\sigma$ to re-weight the error matrix $\mathbf{E}$ defined in  Eq. \ref{eq:sr_error}, then produce the sampling error of each module $M_k$. Here, $(i, j)$ indicates the coordinate of a pixel in module $M_k$.
\begin{align}
\widetilde{E}_{M_k} = \sum_{(i, j) \in M_k} \mathbf{G}_\sigma(i, j) \cdot \mathbf{E}(i, j),
\end{align}

Moreover, at the module level, we define the function $\phi$ to determine whether a module $\mathbf{x}_{M_k}$ in image $\mathbf{x}$ should be updated during each iterative refinement step according to the simulated scanning result, allowing our SRL has the early-stopping mechanism. Then, we calculate the average over $N$ modules in summation. Our Scanning Robust Loss (SRL) can be formulated as
\begin{align}
\mathcal{L}_\text{SR}(\mathbf{x}, \mathbf{y}) = \frac{1}{N}\sum_{k=1}^{N} \phi (\operatorname{sg}[\mathbf{x}_{M_k}], \mathbf{y}_{M_k}) \cdot \widetilde{E}_{M_k},
\end{align}
where $\phi(\cdot, \cdot)$
is a function indicating the update decision. Since $\mathbf{x}_{M_k}$ is not engaged in gradient computing, we introduce stop gradient operator $\operatorname{sg}[\cdot]$ to prevent directly updating $\mathbf{x}_{M_k}$ with respect to $\phi(\mathbf{x}_{M_k}, \mathbf{y}_{M_k})$.

\vspace{-10pt}

\subsubsection{Early-stopping Mechanism.}
In the following, we further elucidate the design of $\phi(\mathbf{x}_{M_k}, \mathbf{y}_{M_k})$. According to the process of QR code scanning, scanners usually consider the average of pixels centered at each module as the representative value of that module. We divide each module to form a matrix of $3 \times 3$ sub-modules, calculate the center sub-module average, and then binarize it using the threshold $\frac{1}{2}$. If the module representative value is identical to our target, then return $0$, early stop it; Otherwise, $1$, implying pixels inside the module should be further refined until correctly matched.
Formally, we define $\phi$ as:
\begin{align}
    \phi(\mathbf{x}_{M_k}, \mathbf{y}_{M_k})=
    \begin{cases}
        0, & \bar{\mathbf{x}}_{M_k} = \mathbf{y}_{M_k}^c, \\
        1, & \bar{\mathbf{x}}_{M_k} \neq \mathbf{y}_{M_k}^c,
    \end{cases},
\end{align}
where $\mathbf{y}_{M_k}^c$ represent the center pixel value of the target module, and the center sub-module mean of $\mathbf{x}_{M_k}$ is 
\begin{align}
    \bar{\mathbf{x}}_{M_k} = \mathbb{I}_{[\frac{1}{2}, 1]}
    \left(
    \sum_{(i, j) \in M_k}
    \mathbf{F}(i, j) \cdot
    \mathcal{G}(\mathbf{x}_{M_k}(i, j))
    \right),
\end{align}
where $\mathbb{I}_A(\cdot)$ is the indicator function of set $A$ and the filter $\mathbf{F}$, designed for 
extracting the center sub-module with module size $m$. We define as:
\begin{align}
    \mathbf{F} = \frac{1}{\lceil \frac{m}{3} \rceil^2}
    \left[
    \begin{array}{ccc}
       \mathbf{O} & \mathbf{O} & \mathbf{O} \\
       \mathbf{O} & \mathbf{I}_{\lceil \frac{m}{3} \rceil \times \lceil \frac{m}{3} \rceil} & \mathbf{O} \\
       \mathbf{O} & \mathbf{O} & \mathbf{O}
    \end{array}
    \right]_{m \times m}.
\end{align}

\subsection{One-stage Generation with iterative refinement via Scanning-Robust Guidance}
\label{sec:OneStage}
Classifier guidance utilizes off-the-shelf pre-trained classification models to perform guidance, \cite{liu2023more, yu2023freedom, bansal2024universal} further extends the conditional probability concept to propose the generalized guidance function that achieves plug-and-play property with only diminutive noisy-image-compatible fine-tuning effort.
Inspired by prior works, we use our SRL to measure the consistency between the intermediate-step images and the target images, transforming the loss into our Scanning Robustness Guidance (SRG).

In the sampling stage, we randomly sample the initial latent from the standard normal distribution and feed it to the pre-trained ControlNet conditioned on the target QR code and user prompt, forming an initial state for the iterative refinement.
We denote $\mathbf{e}_p$ as the prompt embedding, $\mathbf{e}_\text{code}$ as the QR code embedding, $\epsilon_\theta(\cdot, \cdot, \cdot, \cdot)$ as the ControlNet and $\bar{\alpha}_t$ as the parameters in sampling process. Then get the approximated clean latent of each time-step $t$ as
\begin{align}
    \mathbf{z}_{0|t}=\frac{1}{\sqrt{\bar{\alpha}_t}}\left(\mathbf{z}_t - \sqrt{1-\bar{\alpha}_t}\epsilon_\theta(\mathbf{z}_t, t, \mathbf{e}_p, \mathbf{e}_\text{code})\right)
\end{align}

To calculate the SRL, we first decode the latent $\mathbf{z}_{0|t}$ to pixel space and get $\mathbf{x}_{0|t}$. We then use SRL to evaluate the scannability of $\mathbf{x}_{0|t}$. Hence we define our guidance function $F_\text{SR}(\cdot, \cdot)$ as
\begin{align}
    F_\text{SR}(\mathbf{z}_t, \mathbf{y})
    &=\lambda \mathcal{L}_\text{SR} \left(\mathcal{D}_{\theta}
    \left(\frac{1}{\sqrt{\bar{\alpha}_t}}\left(\mathbf{z}_t - \sqrt{1-\bar{\alpha}_t}
    \epsilon_\theta(\mathbf{z}_t, t, \mathbf{e}_p, \mathbf{e}_\text{code})\right)\right), \mathbf{y} \right),
\end{align}
where the $\lambda > 0$ controls the strength of SRL, $\mathcal{D}_{\theta}(\cdot)$ be the pre-trained image decoder of ControlNet to convert the latent to the image in the pixel space. Thus, the guided conditional score estimation become
\begin{align}
\hat{\epsilon}_t = \epsilon_\theta(\mathbf{z}_t, t, \mathbf{e}_p, \mathbf{e}_\text{code})+
\sqrt{1-\bar{\alpha}_t}\nabla_{\mathbf{z}_t}
F_\text{SR}(\mathbf{z}_t, \mathbf{y}).
\end{align}

Followed by DDIM \cite{song2020denoising} sampling process, we sample $\mathbf{z}_{t-1}$ from $q(\mathbf{z}_{t-1} |\mathbf{z}_t, \hat{\mathbf{z}}_{0|t})$. Then we have
\begin{align}
    \mathbf{z}_{t-1} =
    \sqrt{\frac{\bar{\alpha}_{t-1}}{\bar{\alpha}_t}} \left(\mathbf{z}_t-\sqrt{1-\bar{\alpha}_t} \hat{\epsilon}_t \right)
    + \sqrt{1-\bar{\alpha}_{t-1}} \hat{\epsilon}_t.
\end{align}

Finally, We summarize our proposed one-stage QR code generation with iteration refinement in Algo. 1 as presented in supplementary materials.

\subsection{Two-stage Generation with Scanning-Robust Perceptual Guidance}
\label{sec:TwoStage}
\begin{figure}[t]
    \centering
    \includegraphics[height=4cm]{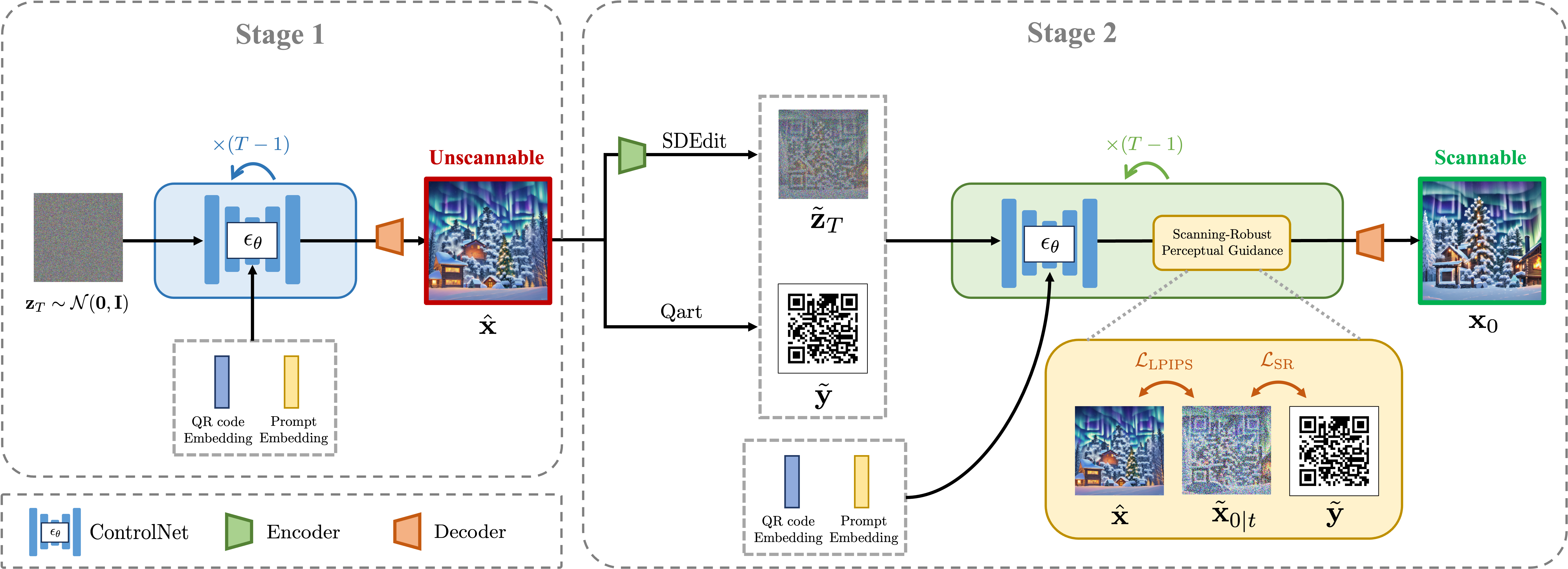}
    \caption{\textbf{An overview of our proposed two-stage generation pipeline with Scanning-Robust Perceptual Guidance (SRPG).} In Stage 1, we utilize the pre-trained plain ControlNet to generate an aesthetic yet unscannable sub-optimal QR code; In Stage 2, we first perform SDEdit to convert the sub-optimal QR code to latent space, then leverage Qart to merge with the target QR code, finally, we apply our proposed iterative refinement to produce aesthetic and scannable QR code.}
    \label{fig:2_stage_pipeline}
    \vspace{-10pt}
\end{figure}

Incorporating SRG with the pre-trained ControlNet as the one-stage generating pipeline generally poses mild deficiencies in aesthetic criteria compared to plain ControlNet. To address this problem, we devise a two-stage pipeline (Fig.~\ref{fig:2_stage_pipeline}) to facilitate the aesthetic property without compromising scannability. In Stage 1, we first generate the preliminary QR code with uncertain scannability $\hat{\mathbf{x}}$ as the reference image. Then, we use the VAE encoder and SDEdit \cite{meng2021sdedit} to convert $\hat{\mathbf{x}}$ to latent space, forming the initial latent noise $\Tilde{\mathbf{z}}_t$ for Stage 2. In Stage 2, we adopt the Qart \cite{qartcodes} algorithm to blend the reference image $\hat{\mathbf{x}}$ with the target QR code, producing the baseline QR code $\Tilde{\mathbf{y}}$. The motivation behind this design is to let ControlNet focus on deficiencies introduced in Stage 1, allowing more ease for ControlNet generation in Stage 2. Then, we apply our iterative refinement introduced in Sec. \ref{sec:OneStage} to generate new $\mathbf{x}_0$. Concerning the balance between scannability and aesthetics, we introduce the Learned Perceptual Image Patch Similarity (LPIPS) loss \cite{zhang2018unreasonable} to bolster the perceptual similarity between the approximated $\Tilde{\mathbf{x}}_{0|t}$ and the reference image $\hat{\mathbf{x}}$ during each sampling step. Finally, we define the Scanning-Robust Perceptual Guidance (SRPG) which both ensures scannability and aesthetics criteria as follows:
\begin{align}
    F_\text{SRP}(\Tilde{\mathbf{z}}_t, \Tilde{\mathbf{y}}, \hat{\mathbf{x}})
    &=\lambda_1 \mathcal{L}_\text{SR}(\Tilde{\mathbf{x}}_{0|t}, \Tilde{\mathbf{y}}) + \lambda_2 \mathcal{L}_\text{LPIPS}(\Tilde{\mathbf{x}}_{0|t}, \hat{\mathbf{x}}),
\end{align}
where coefficients $\lambda_1, \lambda_2$ are hyper-parameters for balancing these two objectives.

\subsection{Post-processing via Scanning-Robust Projected Gradient Descent (SRPGD)}
\label{sec:SRPGD}
In the practical scenarios, if scannability is prioritized and minor blemishes are acceptable, we can further introduce post-processing techniques to ensure higher scannability. Specifically, we use the Projected Gradient Descent (PGD) constraint on the pixel space $\mathcal{P}$ via our proposed Scanning-Robust Loss $\mathcal{L}_\text{SR}$ to optimize the generated QR code until scannability convergence.

\section{Experiment}

\subsection{Comparison with Others Aesthetic QR Code Methods}

\begin{figure}[t]
    \centering
    \includegraphics[height=6.5cm]{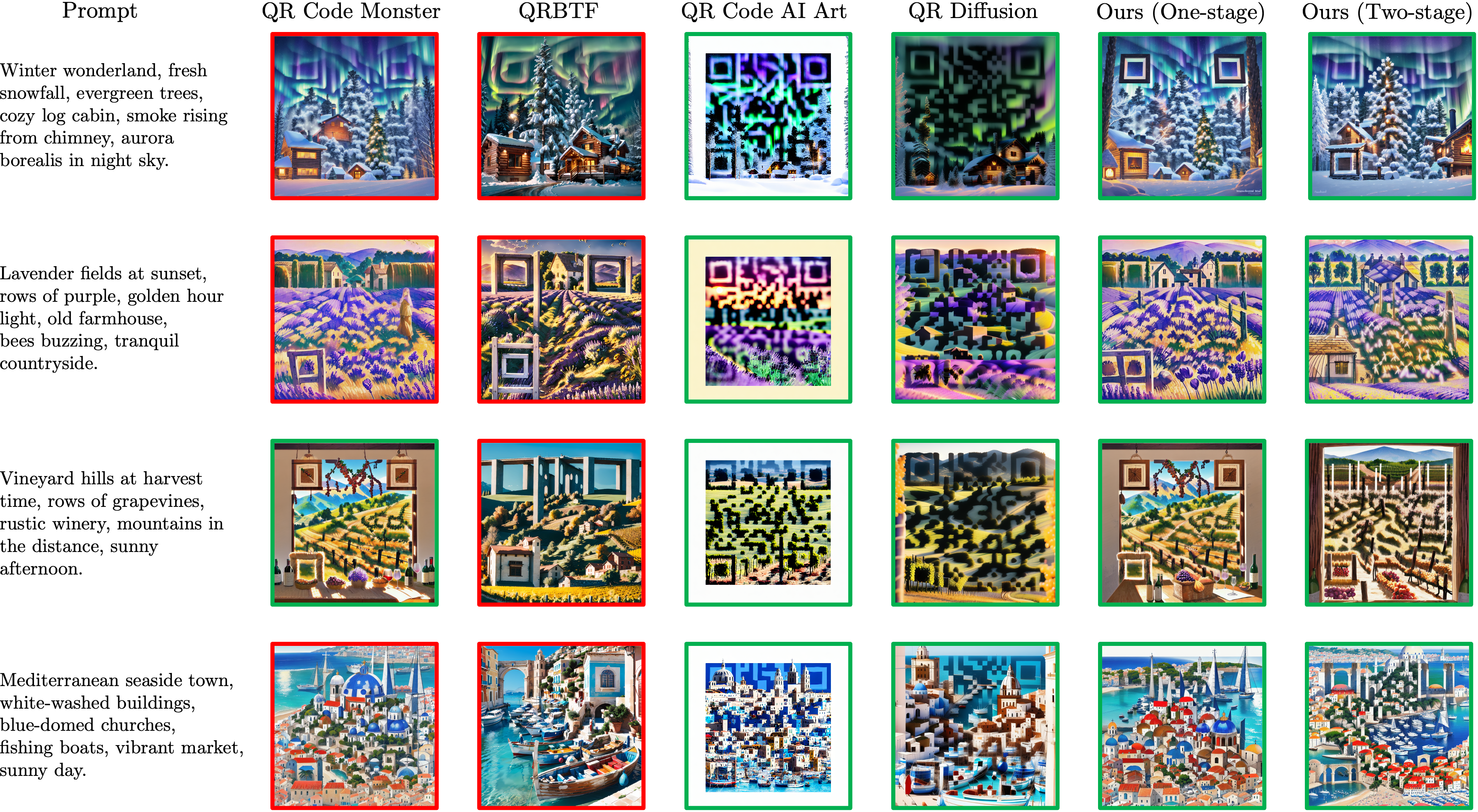}
    \caption{Comparisons with generative-based methods. The green box represents scannable images, while the red box indicates images that cannot be scanned.}
    \label{fig:comparison}
    \vspace{-10pt}
\end{figure}

\subsubsection{Implementation Details.}
We generate 30 text prompts as LDM conditions via GPT-4 for all the comparisons. We use \texttt{Cetus-Mix Whalefall} \footnote{\url{https://huggingface.co/fp16-guy/Cetus-Mix_Whalefall_fp16_cleaned}} as Stable Diffusion checkpoint and QR code Monster v2 \cite{qrcodemonster2023} as the ControlNet \cite{zhang2023adding} checkpoint.
We compared pre-existing methods quantitatively and qualitatively. Proprietary models are accessed via their web apps with their recommended best parameters. Our negative prompt is fixed as \texttt{easynegative}. We utilize a Version 3 QR code, applying mask pattern 4 and padding of 80 pixels. Each module is composed of $20 \times 20$ pixels. The phrase \texttt{Thanks reviewer!} is encoded into our QR code. We set our inference step count to 40 and the ControlNet scale to 1.35. All the experiments are conducted with a single NVIDIA RTX4090 GPU.

\vspace{-10pt}

\subsubsection{Evaluation Metrics.}
We use the iPhone 13's built-in QR code scanner to assess the Scanning Success Rate (SSR) of aesthetic QR codes. We utilize LAION aesthetic predictor \footnote{\url{https://github.com/LAION-AI/aesthetic-predictor}} for quantitative evaluation of aesthetics. LAION aesthetic predictor is an estimator trained on images scored by humans. It predicts a score from 1 (the lowest) to 10 (the highest) to show the images' quality and visual appeal, we referred to the score as the LAION Aesthetic Score (LAS).
Since we have the original target QR codes as ground truths for QR code decoding fidelity evaluations, we can convert our aesthetic QR codes to binary images and compute their module mismatch percentage directly.

\vspace{-10pt}

\subsubsection{Result.}
\begin{table}[t]
    \centering
    \caption{Quantitative results of generative-based methods and our proposed pipeline. Improvements marked in green are compared with QR Code Monster.}
    \vspace{-10pt}
    \label{tab:other_methods}
    \begin{tabular}{lccc}
    \toprule[1pt]
        Method             & LAS $\uparrow$ & SSR $\uparrow$ & \\
        \hline
        QR Diffusion \cite{qrdiffusion} & 6.0611 & 93.33\% \\
        QR Code AI Art \cite{qrcodeaiart2023} & 5.2033 & 76.67\%  \\
        QRBTF \cite{qrbtf2023} & 7.1802 & 36.67\%  \\
        QR Code Monster \cite{qrcodemonster2023} & 7.3369 & 46.67\%  \\
        \hline
        Ours (One-stage + \cite{qrcodemonster2023}) & 6.9164 & 83.33\% &
        \textbf{\textcolor{Green}{(+36.66\%)}} \\
        \rowcolor{cyan!10}
        Ours (Two-stage + \cite{qrcodemonster2023}) & 7.1762 & 86.67\% &
        \textbf{\textcolor{Green}{(+40.00\%)}} \\
        Ours (One-stage + \cite{qrcodemonster2023} + SRPGD) & 6.9007 & 100.00\% &
        \textbf{\textcolor{Green}{(+53.33\%)}} \\
        \rowcolor{cyan!10}
        Ours (Two-stage + \cite{qrcodemonster2023} + SRPGD) & 7.1385 & 96.67\% &
        \textbf{\textcolor{Green}{(+50.00\%)}} \\
    \bottomrule[1pt]
    \end{tabular}
    \vspace{-10pt}
\end{table}

As shown in Tab. \ref{tab:other_methods}, we present the quantitative results of our method in comparison with other generative AI methods, the improvements marked in green are comparisons to QR Code Monster as our baseline. Compared to QRBTF and QR Code Monster, the results indicated that our method achieves a higher SSR compared to QR Diffusion and QR Code AI Art, and our method exhibits greater aesthetic appeal. The images generated by our approach are displayed in Fig. \ref{fig:comparison}. Results show that our one-stage pipeline achieved 83.33\% (+36.66\%) SSR with a 0.4 decrease in LAS; our two-stage pipeline achieved 86.67\% (+40\%) SSR with a 0.16 decrease in LAS. Combined with SRPGD, our one-stage pipeline achieved 100\% (+53.33\%) SSR with a 0.4 decrease in LAS; our two-stage pipeline achieved 96.67\% (+50\%) SSR with a 0.2 decrease in LAS.

\subsection{Ablation Study}
\subsubsection{Influences on Different Scanning-Robust (Perceptual) Guidance Weights.}
As shown in Tab. \ref{tab:one_two_stage}, in the one-stage generation pipeline, as SRG weight increases, the SSR significantly improves with a minor decrease in LAS. In the two-stage generation pipeline, we evaluate the effectiveness of LPIPS perceptual guidance weight $\lambda_2$, we choose $\lambda_1 = 500$ for better comparison. The results showed that LAS improves while SSR preserves as the $\lambda_2$ increases.

\vspace{-10pt}

\subsubsection{Comparison between One-stage and Two-stage Generation.}
\begin{table}[t]
    \centering
    \caption{Quantitative comparisons of one-stage and two-stage pipelines with different SRG or SRPG weights.}
    \label{tab:one_two_stage}
    \vspace{-10pt}
    \begin{tabular}{lcccc}
    \toprule[1pt]
        Method & SRPGD & LAS $\uparrow$ & SSR $\uparrow$ & \\
        \hline
        QR Code Monster \cite{qrcodemonster2023} & & 7.3369 & 46.67\%  \\
        \hline
        One-stage ($\lambda = 400$) & & 6.9116 & 73.33\% & \textbf{\textcolor{Green}{(+26.66\%)}} \\
        One-stage ($\lambda = 500$) & & 6.9164 & 83.33\% &
        \textbf{\textcolor{Green}{(+33.33\%)}} \\
        One-stage ($\lambda = 600$) & & 6.8702 & 80.00\% &
        \textbf{\textcolor{Green}{(+36.66\%)}} \\
        One-stage ($\lambda = 1000$) & & 6.7483 & 90.00\% &
        \textbf{\textcolor{Green}{(+43.33\%)}} \\
        \hline
        Two-stage ($\lambda_1 = 500, \lambda_2 = 2$) & & 7.1248 & 86.67\% &
        \textbf{\textcolor{Green}{(+40.00\%)}} \\
        Two-stage ($\lambda_1 = 500, \lambda_2 = 3$) & & 7.1762 & 86.67\% &
        \textbf{\textcolor{Green}{(+40.00\%)}} \\
        Two-stage ($\lambda_1 = 500, \lambda_2 = 5$) & & 7.1598 & 83.33\% &
        \textbf{\textcolor{Green}{(+40.00\%)}} \\
        Two-stage ($\lambda_1 = 500, \lambda_2 = 10$) & & 7.2011 & 83.33\% &
        \textbf{\textcolor{Green}{(+36.66\%)}} \\
        \hline
        One-stage ($\lambda = 400$) & \Checkmark & 6.8936 & 100.00\% & \textbf{\textcolor{Green}{(+53.33\%)}} \\
        One-stage ($\lambda = 500$) & \Checkmark & 6.9007 & 100.00\% & \textbf{\textcolor{Green}{(+53.33\%)}} \\
        One-stage ($\lambda = 600$) & \Checkmark & 6.8436 & 100.00\% & \textbf{\textcolor{Green}{(+53.33\%)}} \\
        One-stage ($\lambda = 1000$) & \Checkmark & 6.7301 & 100.00\% & \textbf{\textcolor{Green}{(+53.33\%)}} \\
        \hline
        Two-stage ($\lambda_1 = 500, \lambda_2 = 2$) & \Checkmark & 7.0835 & 96.67\%
        & \textbf{\textcolor{Green}{(+50.00\%)}} \\
        Two-stage ($\lambda_1 = 500, \lambda_2 = 3$) & \Checkmark & 7.1385 & 96.67\% & \textbf{\textcolor{Green}{(+50.00\%)}} \\
        Two-stage ($\lambda_1 = 500, \lambda_2 = 5$) & \Checkmark & 7.1260 & 96.67\% & \textbf{\textcolor{Green}{(+50.00\%)}} \\
        Two-stage ($\lambda_1 = 500, \lambda_2 = 10$) & \Checkmark & 7.1490 & 96.67\% & \textbf{\textcolor{Green}{(+50.00\%)}} \\
    \bottomrule[1pt]
    \end{tabular}
\end{table}

As shown in Tab. \ref{tab:one_two_stage}, our proposed two-stage generation pipeline exhibits a higher SSR than our one-stage pipeline while preserving LAS. These results suggested the effectiveness of our proposed two-stage pipeline that further improves the scannability with little cost in aesthetics.

\vspace{-10pt}

\subsubsection{Effectiveness of Scanning-Robust Projected Gradient Descent.}
\begin{figure}[t]
    \centering
    \includegraphics[height=3cm]{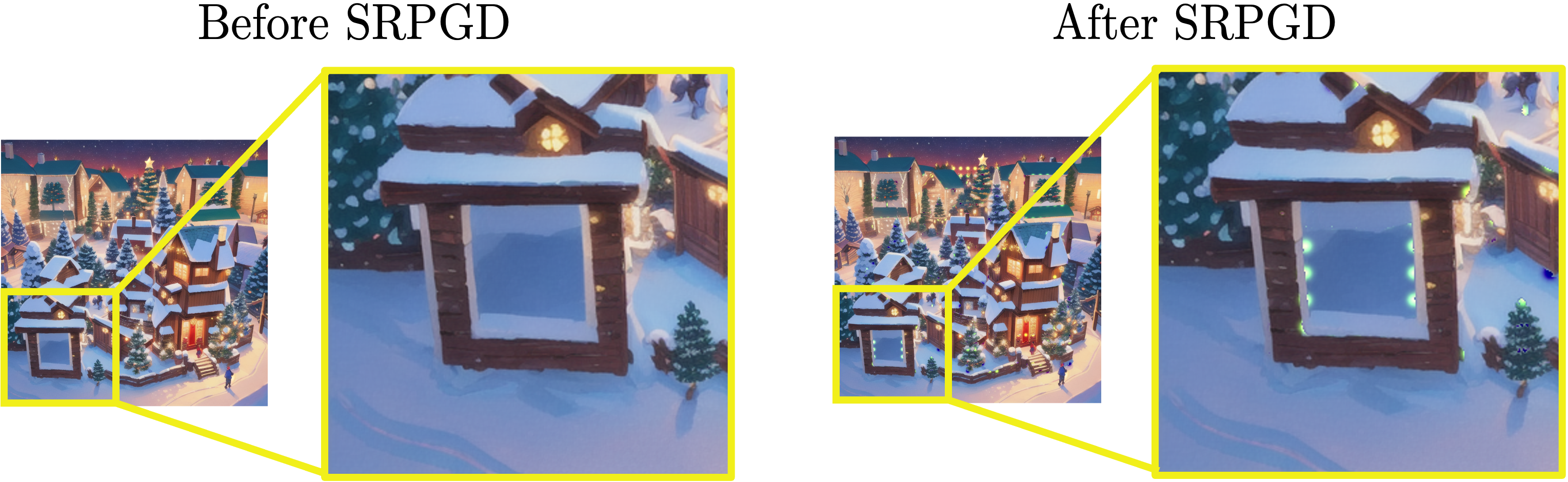}
    \caption{Before-and-after comparison of our proposed SRPGD post-processing method.}
    \label{fig:ablation_study_pgdm}
    \vspace{-10pt}
\end{figure}

Refer to Tab. \ref{tab:one_two_stage}, we can show the effectiveness of our SRPGD post-processing method. Notably, in the one-stage generation pipeline, our post-processing ensures a 100\% SSR while the two-stage generation pipeline still ensures a 96.67\% SSR. All the post-processing variations resulted in diminutive reductions in terms of LAION Aesthetic Score (LAS). Fig. \ref{fig:ablation_study_pgdm} shows qualitative comparisons with enlarged views.

\subsection{Error Analysis}
\subsubsection{Influences on QR Code Error and Score Magnitude with Different Scanning-Robust Guidance Weights.}

\begin{figure}[t]
    \centering
    \begin{subfigure}[b]{0.49\textwidth}
        \includegraphics[width=\textwidth]{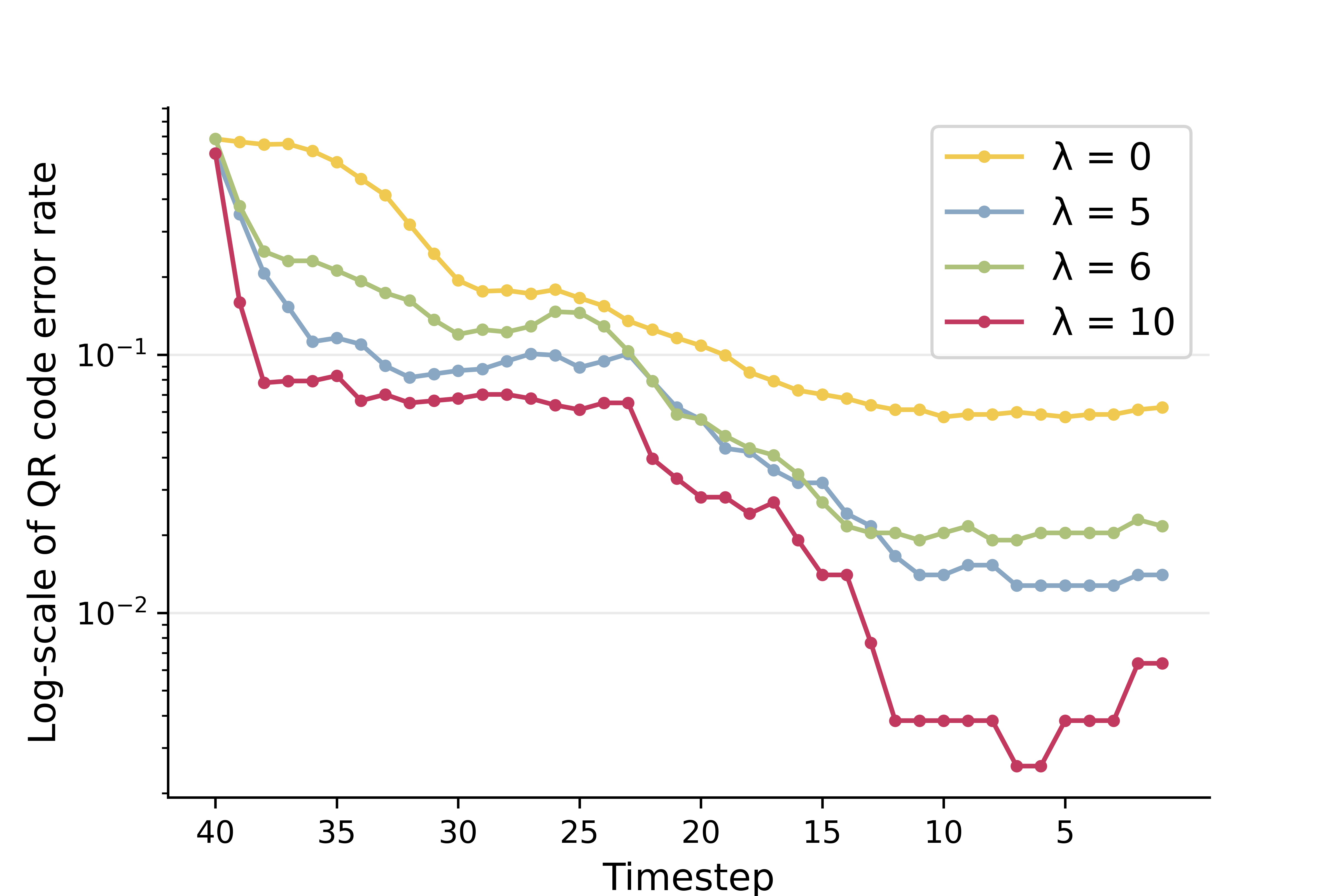}
        \caption{QR code error rate.}
    \end{subfigure}
    \begin{subfigure}[b]{0.49\textwidth}
        \includegraphics[width=\textwidth]{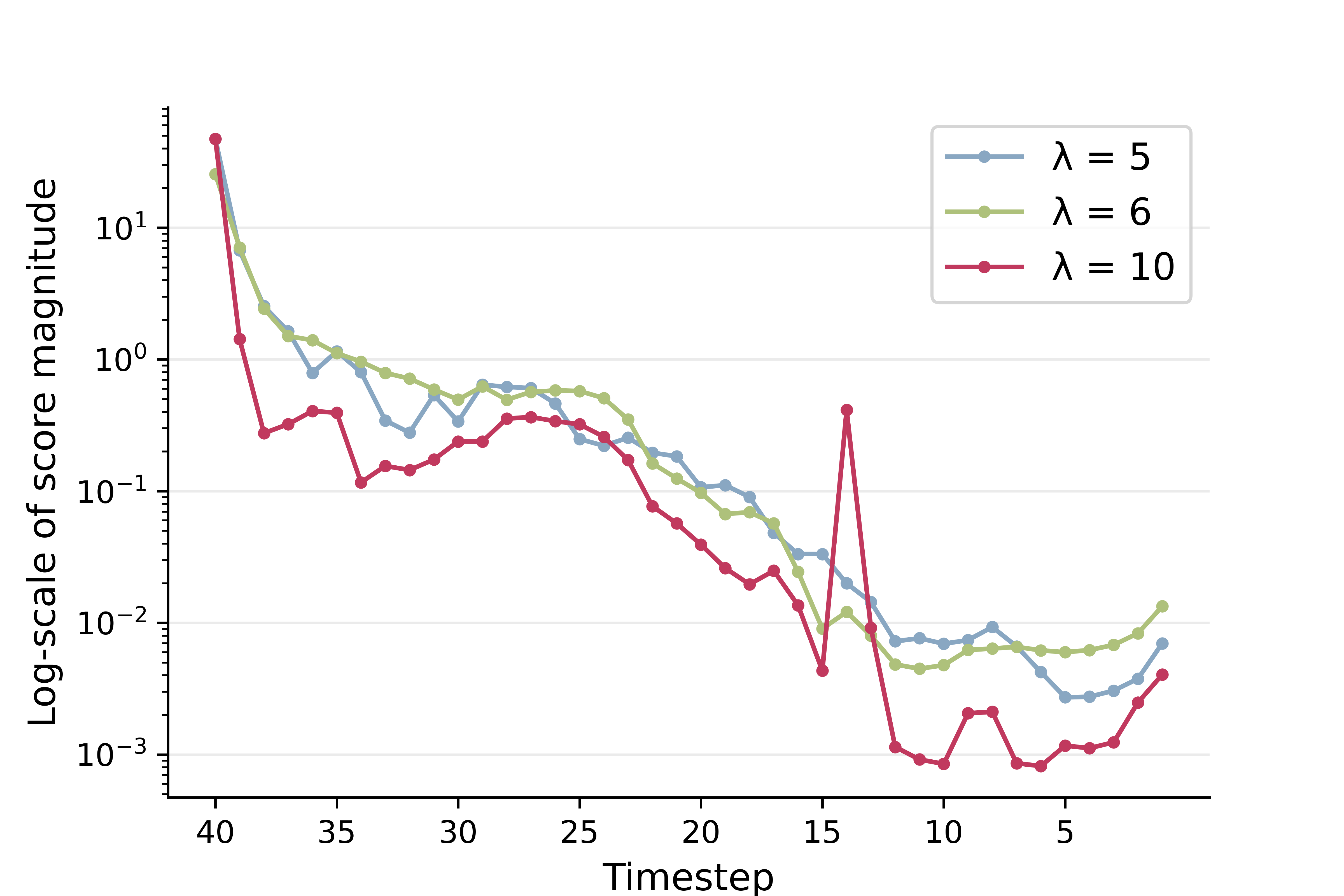}
        \caption{Score magnitude.}
    \end{subfigure}
    \caption{QR code error rate and score magnitude  $\|\nabla_{\mathbf{z}_t} F(\mathbf{z}_t, \mathbf{y})\|_F$ of iterations using our proposed iterative refinement algorithm.}
    \label{fig:qrcode_error}
\end{figure}

As shown in Fig. \ref{fig:qrcode_error} (a), we compare the error rates of a sample with different SRG weights $\lambda$ during iterative refinement steps. We observed that the error plunges within the first 5 iterations with SRG, whereas without SRG, i.e., $\lambda = 0$, decreases gently.
Furthermore, we analyze the change in score magnitude \(\|\nabla_{\mathbf{z}_t} F(\mathbf{z}_t, \mathbf{y})\|_F\) of different SRG weights. We found that the score magnitude decreased over the iterations, indicating the guidance effects diminished over time. This trend is depicted in Fig. \ref{fig:qrcode_error} (b).

\vspace{-10pt}

\subsubsection{Visualization of QR Code Error with Different Timesteps.}

\begin{figure}[t]
    \centering
    \includegraphics[height=2.5cm]{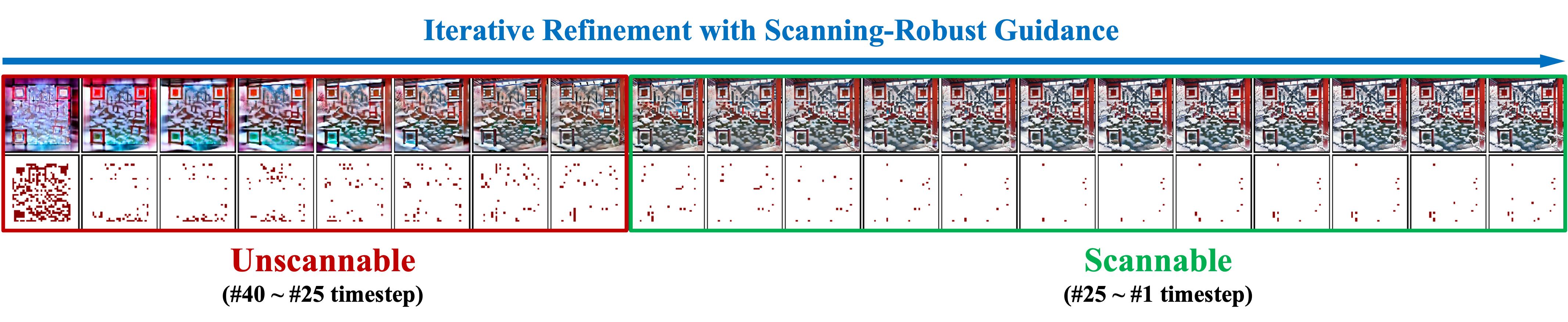}
    \caption{Visualization of $\mathbf{x}_{0 \mid t}$ and its mismatched modules during sampling steps.}
    \label{fig:iterative_refinement_process}
    \vspace{-10pt}
\end{figure}

We visualize different $\mathbf{x}_{0 \mid t}$ images and their corresponding mismatched modules during each sampling step (Fig.~\ref{fig:iterative_refinement_process}). The mismatched modules are marked in red, indicating the inconsistencies between 
scanner-decoded $\mathbf{x}_{0 \mid t}$ and the target QR code $\mathbf{y}$, Initially, $\mathbf{x}_{0 \mid t}$ contains plethora mismatched modules, leading to the unscannable situation. However, the number of mismatched modules significantly decreases as the sampling process proceeds. Moreover, we can observe that the amount of mismatched modules plunges after certain sampling steps. This indicates that the mismatch rate falls within the QR code error-correction capacity, allowing the control reverting to the diffusion model to generate more appealing results.

\vspace{-10pt}

\subsubsection{Visualization of QR Code Module Error.}
\begin{figure}[t]
    \centering
    \includegraphics[height=3.5cm]{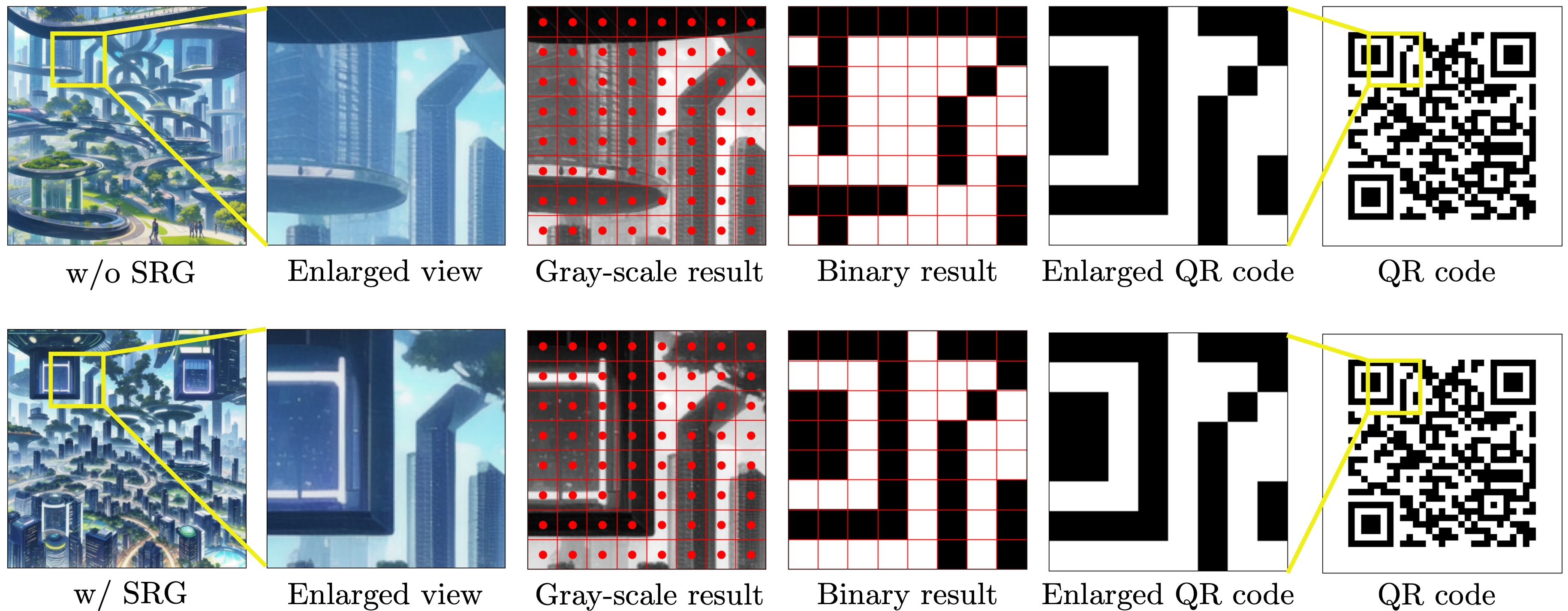}
    \caption{Visual illustration of error analysis.}
    \label{fig:error_analysis}
    \vspace{-10pt}
\end{figure}

We analyze the robustness of the generated results through error analysis. According to Sec. \ref{sec:SRL}, the scanning robustness can be maintained as long as the modules after sampling and binarization yield identical results as the target QR code regardless of pixel color changes within the modules.
Fig. \ref{fig:error_analysis} indicates that our aesthetic QR codes exhibit irregular colors and shapes in their modules. Despite undergoing sampling and binarization, the module results remain consistent with the original QR code. This suggests that our aesthetic QR codes are robust and readable by a standard QR code scanner.

\subsection{User Study}
\begin{table}[t]
    \centering
    \caption{Aesthetic average rank.}
    \label{tab:user_study_aesthetic}
    \vspace{-10pt}
    \centering
    \begin{tabular}{lccc}
         \toprule
            Methods & \textbf{QR Code Monster} \cite{qrcodemonster2023} & \textbf{One-stage} & \textbf{Two-stage} \\
            \midrule
            Avg. Rank & 2.08 & \textbf{1.88} & 2.02 \\
        \bottomrule
    \end{tabular}
    \vspace{5pt}
    \centering
    \caption{Aesthetic average rank on generative-based methods.}
    \label{tab:user_study_different_methods}
    \vspace{-10pt}
    \begin{tabular}{ccccc}
        \toprule
        Methods & \textbf{QR Diffusion} \cite{qrdiffusion} & \textbf{QR Code AI Art} \cite{qrcodeaiart2023} & \textbf{QRBTF} \cite{qrbtf2023}  & \textbf{Ours (Two-stage)} \\
        \midrule
        Avg. rank & 3.18 & 2.71 & \textbf{1.86} & 2.25 \\
        \bottomrule
      \end{tabular}
    \vspace{-10pt}
\end{table}

To further evaluate the effectiveness of our proposed approach, we conducted a user-subjective survey with 387 participants.

\vspace{-10pt}

\subsubsection{Influences on Visual Quality with Scanning-Robust Guidance.}
Refer to Tab. \ref{tab:user_study_aesthetic}, most participants think our one-stage pipeline exhibits the most visual appeal; our two-stage pipeline is second place, and methods without SRG are the least visually appealing. These results demonstrate that our approach causes little degradation in terms of visual quality. Moreover, there are signs of further improvement.

\vspace{-10pt}

\subsubsection{Comparison of Visual Quality with Different Generative-based Methods.}
Participants are further asked to rate images generated by pre-existing methods. Refer to Tab. \ref{tab:user_study_different_methods}, while QRBTF \cite{qrbtf2023} achieved first place, our approach closely follows, indicating a marginal difference. Refer to Tab. \ref{tab:other_methods}, considering the QRBTF limited scannability, it's evident that our approach is the premier method for effectively balancing visual attractiveness with scannability.

\vspace{-10pt}

\subsubsection{Impacts on Visual Quality of Scanning-Robust Projection Gradient Descent.}
To further assess the negative impact on visual quality in practical scenarios, the user-subjective survey shows 92\% participants think the artifacts after post-processing are acceptable.

\section{Conclusion}
In this paper, we introduce two novel diffusion-based pipelines for aesthetic QR code generation. We develop the Scanning-Robust Loss (SRL) to enhance the scannability of QR codes and have established a bridge between the SRL and Scanning-Robust Guidance (SRG). Moreover, we introduce convex-optimization-based Scanning-Robust Projected Gradient Descent (SRPGD) to ensure scannability convergence.
Compared to other pre-existing methods, our approach enhances the scanning success rate without significantly compromising visual quality. The QR codes generated through our approach are capable of applications in real-world scenarios.

\subsubsection*{Acknowledgements.}
We genuinely thank Ernie Chu for the inspirational discussions with him and for his insightful suggestions and feedback. In addition, we thank Steven Wu for his help explaining some previous works.

\clearpage
\bibliographystyle{splncs04}
\bibliography{main}

\end{document}